# Experiments of Distance Measurements in a Foliage Plant Retrieval System

Abdul Kadir, Lukito Edi Nugroho, Adhi Susanto and Paulus Insap Santosa

*Gadjah Mada University, Indonesia*
{akadir, lukito, insap}@mti.ugm.ac.id, susanto@te.ugm.ac.id

*Abstract*

*One of important components in an image retrieval system is selecting a distance measure to compute rank between two objects. In this paper, several distance measures were researched to implement a foliage plant retrieval system. Sixty kinds of foliage plants with various leaf color and shape were used to test the performance of 7 different kinds of distance measures: city block distance, Euclidean distance, Canberra distance, Bray-Curtis distance, $\chi^2$ statistics, Jensen Shannon divergence and Kullback Leibler divergence. The results show that city block and Euclidean distance measures gave the best performance among the others.*

**Keywords:** *Distance Measure, Retrieval system, City block distance, Euclidean distance*

## 1. Introduction

Several image retrieval systems have been developed with various approaches. One of important components in such systems are how to compute rank/score of the image query with each image in the references (database) that actually incorporated a distance measure such as city block, Euclidean, Canberra distance or Bray-Curtis. For example, Euclidean distance measures were used in [1] and [2]. Canberra distance measure was used in [3]. Meanwhile, city block distance measure was used in [4] and [5]. It was also used by Wang et al. [6] as feature for the first-state leaf image retrieval. The example of Bray-Curtis application is to analyze ecological data [7]. Each distance measure has its complexity. Generally, Euclidean distance is more popular than other distance measures for image retrieval. However, city block distance is more desirable than Euclidean distance because of its simplicity [8].

In this research, seven distance measures: city block distance, Euclidean distance, Canberra distance, Bray Curtis distance, Jensen Shannon divergence and Kullback Leibler divergence, had been explored. The main reason was to investigate which distance measures that will give the best performance in the foliage plant retrieval system.

To implement the retrieval system, several features based on our previous researches [9] [10] and several additional features were used. The system incorporated shape, color, vein and texture features. To improve performance of the system, five kinds of geometric features plus texture features were included. Therefore, shape features were descriptors derived from Polar Fourier Transform, eccentricity, roundness, dispersion, solidity and convexity. Color features incorporated the mean, standard deviation and skewness of colors. Vein features contained two kind of features extracted from vein of the leaf. Texture features contained five features derived from Gray Level Occurrence Matrix (GLCM).

The retrieval system was designed to get the top one, top three, and top five of plants that have most similarity with the leaf of query. If the top one is used, the system will give answer like NN or SVM based system [11], which is the most similar. However, if the result of the top one fails to give the right answer, the top three or the top five could give alternative solutions for users.

The rest of the paper organized as follows: Section 2 describes several features involved in the retrieval system, Section 3 explains seven distance measures, Section 4 describes scheme of the experiments and reports the experimental results and Section 5 concludes the results.

## 2. Leaf Features

### 2.1. Geometric Features

Geometric features are used to capture shape of the leaf. There were five geometric features used in the retrieval system: eccentricity, roundness, dispersion, solidity, and convexity.

Eccentricity or slimness ratio is defined as

$$eccentricity = \frac{w}{l} \tag{1}$$

In this case, w is the length of the leaf' minor axis and l is the length of the leaf' major axis.

Roundness or circularity ratio is defined as

$$roundness = \frac{A}{p^2} \tag{2}$$

where A is the area of the leaf and P is the perimeter of the leaf.

Dispersion is the ratio between the radius of the maximum circle enclosing the region and the minimum circle that can be contained in the region [12]. Formula of dispersion is as below

$$dispersion = \frac{\max(\sqrt{(x_i - \bar{x})^2 + (y_i - \bar{y})^2})}{\min(\sqrt{(x_i - \bar{x})^2 + (y_i - \bar{y})^2})} \tag{3}$$

In the above formula, $(\bar{x}, \bar{y})$ is the centroid of the leaf and $(x_i, y_i)$ is the coordinate of a pixel in the leaf contour.

Solidity and convexity are defined as [13]

$$solidity = \frac{area}{convex\ area} \tag{4}$$

$$convexity = \frac{convex\ perimeter}{perimeter} \tag{5}$$

In both equations, convex represents a convex hull, the smallest convex set containing all points in an object. It is like a rubber band that surrounds the edge of an object. Figure 1 gives an example of a convex hull of the leaf. Actually, the convex hull can be calculated using 'Graham Scan' algorithm [14].

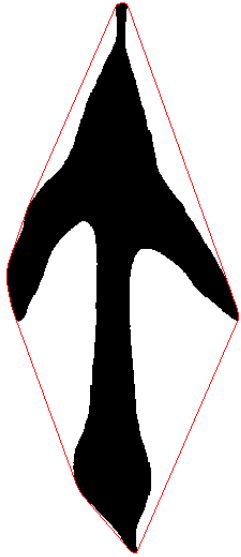

**Figure 1. The Convex Hull for a Shape Region of the Leaf**

### 2.2. Polar Fourier Transform

Proposed by Zhang [8], Polar Fourier Transform (PFT) has properties that are very useful for representing shape of leaves. The descriptors extracted from PFT are called Generic Fourier Descriptors (GFDs). PFT that was used in this research is defined as

$$PFT(\rho,\phi) = \sum_r \sum_i f(\rho,\theta).\exp\left[ j2\pi\left( \frac{r}{R}\rho + \frac{2\pi}{T}\phi \right) \right] \qquad (6)$$

where:

- $0 \leq r < R$ dan $\theta_i = i(2\pi/T)$ $(0 \leq i < T)$; $0 \leq \rho < R$, $0 \leq \phi < T$,
- R is radial frequency resolution,
- T is angular frequency resolution.

Computation of PFT is described as follow. For example, there is an image I = {f(x, y); $0 \leq x < M$, $0 \leq y < N$}. Firstly, the image is converted from Cartesian space to polar space $I_p$ = {f(r,θ); $0 \leq r < R$, $0 \leq \theta < 2\pi$}, where R is the maximum radius from center of the shape. The origin of polar space becomes as center of space to get translation invariant. The centroid ($x_c$, $y_c$) was calculated by using formula

$$x_c = \frac{1}{M}\sum_{i=0}^{M-1} x, \quad y_c = \frac{1}{N}\sum_{i=0}^{N-1} y, \qquad (7)$$

where M is total rows of the image and N is total columns of the image. Whereas, (r, ө) is computed by using:

$$r = \sqrt{(x-x_c)^2 + (y-y_c)^2}, \theta = \arctan\frac{y-y_c}{x-x_c} \qquad (8)$$

Rotation invariance is achieved by ignoring the phase information in the coefficient. Therefore, only the magnitudes of coefficients are retained. Meanwhile, to get the scale invariance, the first magnitude value is normalized by the area of the circle and all the magnitude values are normalized by the magnitude of the first coefficient. So, the GFDs are

$$GFD_s = \left\{\frac{PF(0,0)}{2\pi r^2}, \frac{PF(0,1)}{PF(0,0)}, \dots, \frac{PF(0,n)}{PF(0,0)}, \dots, \frac{PF(m,0)}{PF(0,0)}, \dots, \frac{PF(m,n)}{PF(0,0)}, \right\} \qquad (9)$$

where m is the maximum number of the radial frequencies and n is the maximum number of angular frequencies. In this research, m = 6 and n = 4.

### 2.3. Gray Level Occurrence Matrix

GLCM had been used in several applications such as in [1] and [15] for image retrieval and classification systems. Several features can be extracted from GLCM. GLCM enables to obtain valuable information about the relative position of the neighboring pixels in an image. The co-occurrence matrix GLCM(i, j) counts the co-occurrence of pixels with grey value i and j at given distance d. The direction of neighboring pixels to represents the distance can be selected, for example 135°, 90°, 45°, or 0°, as illustrated in Figure 2. A common choice is to compute GLCMs for a distance of one (i.e., adjacency) and four directions, 0, 45, 90, and 135 degrees.

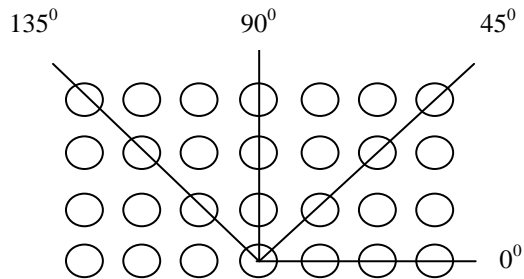

**Figure 2. Direction in obtaining GLCM**

There are many scalar quantities derived from GLCM as proposed by Haralick [16]. However, only few of them have significant contribution in several applications. For example, Newsam & Kamath [17] proposed five GLCMs derived features: angular second moment (ASM), contrast, inverse different moment (IDM), entropy, and correlation.

ASM (or energy) measures textural uniformity (i.e. pixel pairs repetition) [18]. The ASM has the highest value when the distribution of the grey levels constant or periodic. Mathematically, ASM is defined as

$$ASM = \sum_{i=1}^{L}\sum_{j=1}^{L}\left(GLCM(i,j)^2\right) \qquad (10)$$

Contrast measures the coarse texture or variance of the grey level. The contrast is expected to be high in coarse texture, if the grey level of contrast corresponds to large local variation of the grey level [18]. Mathematically, this feature is calculated as

$$Contrast = \sum_{i=1}^{L}\sum_{j=1}^{L}(i-j)^2\left(GLCM(i,j)\right) \qquad (11)$$

The IDM (or homogeneity) measures the local homogeneity a pixel pair. The homogeneity is expected to be large if the grey levels of each pixel pair are similar [19]. This feature is computed as

$$IDM = \sum_{i=1}^{M}\sum_{j=1}^{N}\frac{(GLCM(i,j))^2}{1+(i-j)^2} \qquad (12)$$

Entropy measures the degree of disorder or non-homogeneity of the image. Large values of entropy correspond to uniform GLCM. For texturally uniform image, the entropy is small. It is computed as

$$Entropy = -\sum_{i=1}^{L}\sum_{j=1}^{L}GLCM(i,j) x \log(GLCM(i,j)) \qquad (13)$$

Correlation texture measures the linear dependency of gray levels on those of neighboring pixels. This feature is computed as

$$Correlation = \sum_{i=1}^{L}\sum_{j=1}^{L}\frac{(ij)(GLCM(i,j)-\mu_1'\mu_2')}{\sigma_i'\sigma_j j'} \qquad (14)$$

where

$$\mu_i' = \sum_{i=1}^{L}\sum_{j=1}^{L} i * GLCM(i,j) \qquad (15)$$

$$\mu_j' = \sum_{i=1}^{L}\sum_{j=1}^{L} j * GLCM(i,j) \qquad (16)$$

$$\sigma_i^2 = \sum_{i=1}^{L}\sum_{j=1}^{L} GLCM(i,j)(i-\mu_i')^2 \qquad (17)$$

$$\sigma_j^2 = \sum_{i=1}^{L}\sum_{j=1}^{L} GLCM(i,j)(i-\mu_i')^2 \qquad (18)$$

In fact, GLCM is dependent to rotation. Therefore, to achieve rotational invariant features, the GLCM features corresponding to four directions (135º, 90º, 45º, or 0º) are firstly calculated and then be averaged [20].

### 2.4. Color Features

Color moments represent color features to characterize a color image. For example, it was used in [19] for skin texture recognition. Features can be involved are mean ($\mu$), standard deviation ($\sigma$), and skewness ($\theta$). For RGB color space, the three features are extracted from each plane R, G, and B. The formulas to capture those moments:

$$\mu = \frac{1}{MN} \sum_{i=1}^{M} \sum_{j=1}^{N} P_{ij} \quad (19)$$

$$\sigma = \sqrt{\frac{1}{MN} \sum_{i=1}^{M} \sum_{j=1}^{N} (P_{ij} - \mu)^2} \quad (20)$$

$$\theta = \frac{\sum_{i=1}^{M} \sum_{j=1}^{N} (P_{ij} - \mu)^3}{MN\sigma^3} \quad (21)$$

M and N are the dimension of image. $P_{ij}$ is the values of color on column $i_{th}$ and row $j_{th}$.

### 2.5. Vein Features

Vein features contains two features extracted from vein. Vein is obtained by using morphological opening [20]. That operation is performed on the gray scale image with flat, disk-shaped structuring element of radius 1 and 2 and subtracted remained image by the margin. As a result, a structure like vein is obtained. Figure 3 shows an example of vein resulted by such operation.

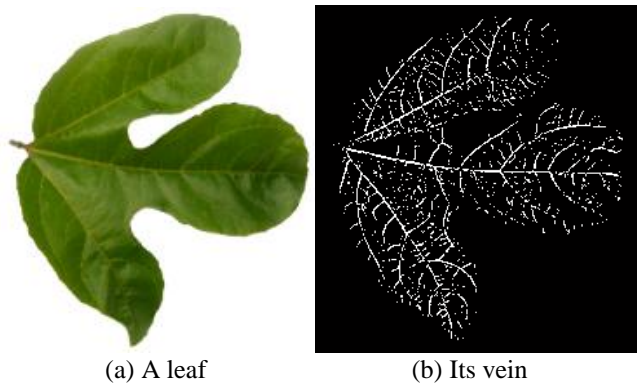

(a) A leaf        (b) Its vein

**Figure 3. Illustration of Vein Processed by using Morphological Operation**

Based on that vein, two features are calculated as follow:

$$V_1 = \frac{A_1}{A}, \quad V_2 = \frac{A_2}{A} \tag{22}$$

In this case, $V_1$ and $V_2$ represent features of the vein, $A_1$ and $A_2$, are total pixels of the vein, and $A$ denotes total pixels on the part of the leaf.

## 3. Distance Measures

There are many distance measures that were used in content-based image retrieval (CBIR) [8, 21]. Seven of them were investigated in this research: city block distance, Euclidean distance, Canberra distance, Bray Curtis distance, $\chi^2$ statistics, Jensen Shannon divergence and Kullback Leibler divergence.

For simplicity of explanation, in seven distance measures will be discussed, d(Q,R) is distance between features of query (Q) and features of a reference (R). Meanwhile, N is number of features.

### 3.1. City Block Distance

The city block distance is the simplest distance measure in computation. It is defined as

$$d(Q, R) = \sum_{i=1}^{N} |Q_i - R_i| \tag{23}$$

This distance is also known as Manhattan distance.

### 3.2. Euclidean Distance

The Euclidean distance is well known and widely used as a distance measure in image retrieval systems. The distance is obtained by using

$$d(Q, R) = \sum_{i=1}^{N} (Q_i - R_i)^2 \tag{24}$$

### 3.3. Canberra Distance

The Canberra distance is often used for data scattered around an origin. The generalized equation is given in the form

$$d(Q,R) = \sum_{i=1}^{N} \frac{|Q_i - R_i|}{|Q_i| + |R_i|} \tag{25}$$

### 3.4. Bray-Curtis Distance

The Bray-Curtis distance is actually not a distance measure (http://en.wikipedia.org/wiki/Bray-Curtis_dissimilarity) because it does not satisfy triangle equality. Therefore, it is sometime called as Bray-Curtis dissimilarity/similarity rather than Bray-Curtis distance. It is defined as

$$d(Q,R) = \frac{\sum_{i=1}^{N} |Q_i - R_i|}{\sum_{i=1}^{N} (Q_i - R_i)} \tag{26}$$

### 3.5. $\chi^2$ statistics

$\chi^2$ statistics is defines as

$$d(Q,R) = \frac{\sum_{i=1}^{N} (Q_i - R_i)^2}{m_i}, \tag{27}$$

where $m_i = \frac{Q_i + R_i}{2}$. This quantity measures how unlikely it is that one distribution was drawn from the population represented by the other [8].

### 3.6. Kullback Leibler Divergence

The Kullback Leibler divergence is defined as [21]

$$d(Q,R) = \sum_{i=1}^{N} Q_i \log \frac{Q_i}{R_i} \tag{28}$$

According the formula, it is not symmetric, because d(**Q**,**R**) is not equal d(**R**,**Q**).

### 3.7. Jensen Shannon Divergence

Jensen Shannon divergence, also called as Jeffrey divergence, is an empirical extension of Kullback Leibler divergence [21]. It is defined as

$$d(Q,R) = \sum_{i=1}^{N} Q_i \log \frac{2Q_i}{Q_i + R_i} + R_i \log \frac{2R_i}{Q_i + R_i} \qquad (29)$$

## 4. Experiments and Results

To investigate the role of each distance measure, a retrieval system has been developed. It used a distance measure for every testing. Experiments were accomplished by using 60 kinds of foliage plants. Sample of plants is shown in Figure 4. Every plant contains 50 leaves for references and 20 leaves for query purpose. However, the actual number of references per plant was varying from 5, 10, 15, 20, 25, 30, 35, 40, 45 and 50.

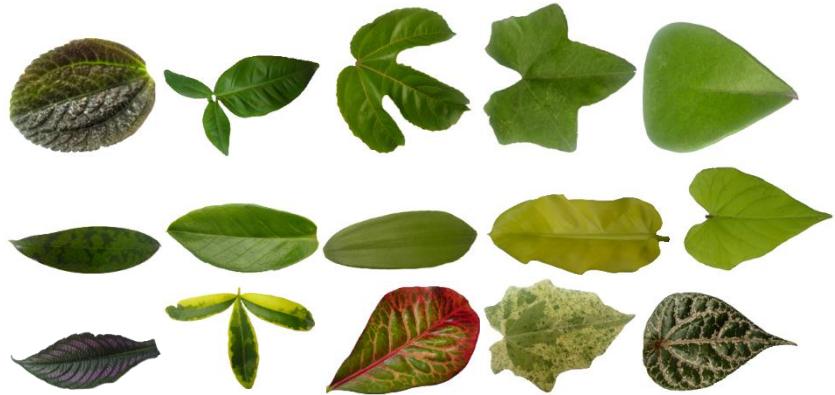

**Figure 4. Sample of Plants used in System' Testing**

The block diagram of the retrieval system is shown in Figure 5. First of all, the leaf of query is inputted to the system. Preprocessing does basic processes such segmenting leaf from its background and converting the RGB image into binary image for next processing. Then, features of the query' leaf are extracted, normalized in range [0, 1] and compared by features of every leaf in the database. In this case, the distance measure is used to compute every rank of the reference of leaf. Based on the ranks of leaves, five plants with smallest score are selected to be outputted.

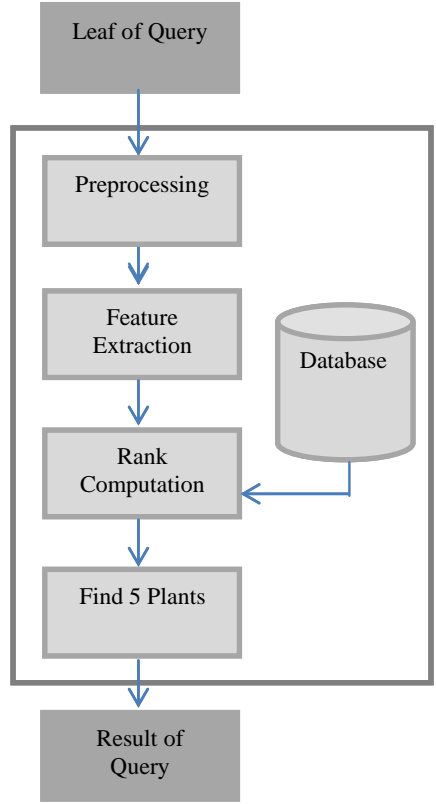

**Figure 5. Foliage Plant Retrieval System**

The basic of rank calculation is accomplished by using a distance measure. However, the actual rank computation is more complex, because of combination of several kinds of features. In fact, there were 4 kinds of features: 1) shape, 2) color, 3) texture and 4) vein features. Each feature is calculated by using a distance measure. After that, the rank is computed by using

$$rank = k_s\ d_s + k_c\ d_c + k_t\ d_t + k_v\ d_v \qquad (30)$$

where $k_s$, $k_c$, $k_t$ and $k_v$ are the weighting coefficient for shape, color, texture, and vein features respectively, $d_s$, $d_c$, $d_t$, and $d_v$ are the Euclidean distance for shape, color, texture, and vein features respectively. In this case, $d_s$ is calculated based on GFDs (Eq. 9), $d_c$ is calculated based on mean ($\mu$), standard deviation ($\sigma$), and skewness ($\theta$) of colors (Eq. 19, 20 and 21), $d_t$ is calculated based on ASM (Eq. 9), contrast (Eq, 10), IDM (Eq. 11), entropy (Eq. 12) and correlation (Eq. 13), and dv is calculated based on $V_1$ and $V_2$ (Eq. 22).

Accuracy of the system is calculated using

$$accuracy = \frac{n_r}{n_t} \qquad (31)$$

where $n_r$ is relevant number of images and $n_t$ is the total number of query. For all experiments, total number of query was 1200.

Other performance measurement called Recall Precision Pair (RPP) [8] is used to illustrate the performance of the system in the other way. The precision and recall are defined as

$$precision = \frac{r}{n_1} = \frac{number\ of\ relevant\ retrieved\ images}{number\ of\ retrieved\ images} \quad (32)$$

$$recall = \frac{r}{n_2} = \frac{number\ of\ relevant\ retrieved\ images}{number\ of\ relevant\ images\ in\ database} \quad (33)$$

First experiments were comparing the performance of every distance measure by using $k_s$, $k_c$, $k_t$ and $k_v$ were equal 0.25, 50 leaves per plant as references, 20 leaves per plant for testing. The results is shown in Table 1 and the graph of its precision-recall is shown in Figure 6. Other results of experiments based on $k_s$= 0.3235 $k_c$= 0.4245 $k_t$ = 0.1059 and $k_v$ = 0.1471 is visualized in Figure 7. The figure shows that the city block and the Euclidean distance give almost same performance and they are the best. Performance of the system that used Canberra distance is lower than that used the Euclidean distance or city block distance. However, it is better than the others. $\chi^2$ statistics and Jensen Shannon divergence give almost same performance. Meanwhile, Kullback Leibler divergence are not good to be involved in the foliage plant retrieval system.

**Table 1. Performance of the System for Various Distance Measures using $k_s$, $k_c$, $k_t$ and $k_v$ = 0.25 and references = 50**

| No. | Measurement | Performance (%) | | | Time (seconds) |
|---|---|---|---|---|---|
| | | Top 1 | Top 3 | Top 5 | |
| 1 | City block distance | 90.0833 | 97.1667 | 98.7500 | 282 |
| 2 | Euclidean distance | 89.3333 | 96.6667 | 98.9167 | 333 |
| 3 | Canberra distance | 87.5000 | 96.6667 | 97.9167 | 315 |
| 4 | Bray-Curtis distance | 85.1167 | 95.0833 | 97.2500 | 322 |
| 5 | $\chi^2$ statistics | 84.7500 | 95.6667 | 97.9167 | 360 |
| 6 | Jensen Shannon Divergence | 84.4167 | 95.9167 | 98.0000 | 400 |
| 7 | Kullback Leibler Divergence | 21.2500 | 43.4167 | 57.5000 | 283 |

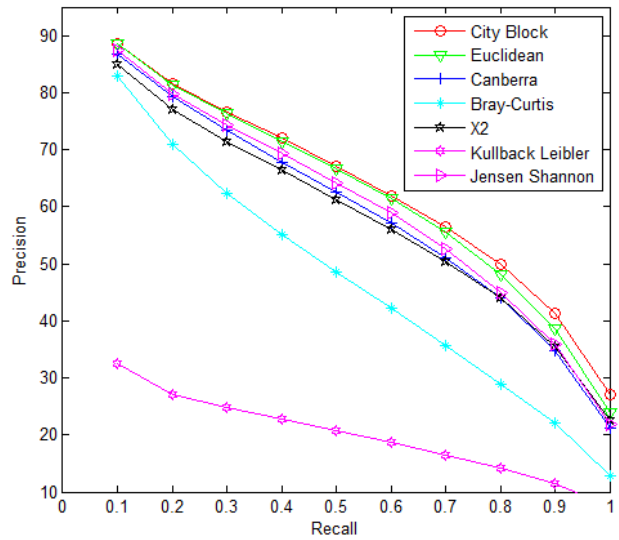

**Figure 6. Average Precision-recall of 1200 Retrievals**

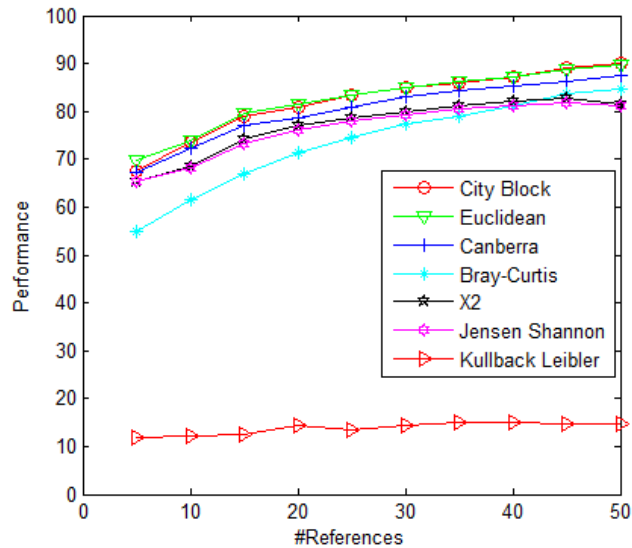

**Figure 7. Performance of Seven Distance Measures**
**$k_s$= 0.3235 $k_c$= 0.4245 $k_t$ = 0.1059 and $k_v$ = 0.1471**

Several experiments in selecting parameters $k_s$, $k_c$, $k_t$ and $k_v$ had been done. The results show that the city block distance or the Euclidean distance are good to be selected as a distance measure in the foliage plant retrieval system. For more than 35 references, the city block distance gave better performance than Euclidean one. Of course, this result give a benefit in time efficiency. Based on time required, the computation using city block distance is faster than using Euclidean distance.

The optimum performance was reached by using $k_s$= 0.1612 $k_c$= 0.4839 $k_t$ = 0.1936 and $k_v$ = 0.1613 when block city distance was used. In this case, the results were 91.9167% for top 1, 97.4167% for top 3 and 98.7500% for top 5. Those results are better than in our previous research [10] (that gave 90.8% of accuracy for top 1).

Besides using our dataset, this research also tried to apply dataset Flavia [20]. By using city block distance measure and parameters $k_s$= 0.1612 $k_c$= 0.4839 $k_t$ = 0.1936 and $k_v$ = 0.1613, the system gave the performance 92.8125% of accuracy (top 1). This performance are better than the original performance (90.312% in [20]). It means that the system can handle not only plants with various color but also plants with green color only.

## 5. Conclusion

Selecting distance measure in the foliage retrieval system is an important component should be considered. The experiments shows that by using the mentioned features, the city block and the Euclidean distance are better than other distance measures (Canberra distance, $\chi^2$ statistics, Bray-Curtis distance, Kullback Liebler divergence and Jensen Shannon divergence). By using more than 35 references, the city block distance gave better performance than Euclidean one. However, the distance measure such as Canberra distance or Bray-Curtis distance should be considered when testing using other features.

## References

[1] B. Jyothi, Y. M. Latha and V. S. K. Reedy, "Medical Image Retrieval using Multiple Features", Advances in Computational Sciences and Technology, vol. 3, no. 3, **(2010)**, pp. 387-396.
[2] A. Khaparde, N. Jain, S. Mantha and N. S. Chowdary, "Searching Query By Color Content of An Image Using Independent Component Analysis", International journal of Enterprise and business Systems, vol. 1, no. 2, **(2011)**.
[3] P. S. Hiremath and J. Pujari, "Content Based Image retrieval based on Color, Texture and Shape features using Image and Its Complement", International Journal of Computer Science and Security, vol. 1, no. 4, **(2010)**, pp. 25-35.
[4] S. M. Lee, H. J. Bae and S. H. Jung, "Efficient Content-Based Image retrieval Methods using Color and Texture", ETRI Journal, vol. 20, no. 3, (1998), pp. 272-283.
[5] B. G. Prasad, S. K. Gupta and K. K. Biswas, "Color and Shape Index for region-Based Image Retrieval", LNCS 2029, **(2001)**, pp. 716-725.
[6] Z. Wang, Z. Chi, D. Feng and Q. Wang, "Leaf Image Retrieval with Shape Features", LNCS 1929, **(2000)**, pp. 477-487.
[7] S. C. Gosslee and D. L. Urban, "The Ecodist Package for Dissimilarity-based Analysis of Ecological Data", Journal of Statistical Software, vol. 22, no. 7, **(2007)**.
[8] D. Zhang, "Image Retrieval Based on Shape", Unpublished Dissertation, Monash University, **(2002)**.
[9] A. Kadir, L. E. Nugroho, A. Susanto and P. I. Santosa, "A Comparative Experiment of Several Shape Methods in Recognizing Plants", International Journal of Computer Science & Information Technology, vol. 3, no. 3, **(2011)**, pp. 256-263.
[10] A. Kadir, L. E. Nugroho, A. Susanto and P. I. Santosa, "Foliage Plant Retrieval Using Polar Fourier Transform, Color Moments and Vein Features", Signal & Image Processing: An International Journal, vol. 2, no. 3, **(2011)**, pp. 1-13.
[11] K. Singh, I. Gupta and S. Gupta, "SVM-BDT PNN and Fourier Moment Technique for Classification of Leaf Shape", International Journal of Signal Processing, Image Processing and Pattern Recognition, vol. 3, no. 4, **(2010)**, pp. 67-78.
[12] M. S. Nixon and A. S. Aguado, "Feature Extraction and Image Processing", Newness, Wobun, **(2002)**.
[13] J. C. Russ, "The Image Processing Handbook", CRC Press, Boca Raton, **(2011)**.
[14] M. T. Goodrich, R. Tammasia, "Algortm Design", John Wiley & Sons, **(2002)**.
[15] A. Ehsanirad, "Plant Classification Based on Leaf Recognition", International Journal of Computer Science and Information Security, vol. 8, no. 4, **(2010)**, pp. 78-81.
[16] A. D. Kulkarni, "Artificial Neural Networks for Image Understanding", Van Nostrand Reinhold, New York, **(1994)**.
[17] S. Newsam and C. Kammath, "Comparing Shape and Texture Features for Pattern Recognition in Simulation Data", IS&T/SPIE's Annual Symposium on Electronic Imaging, San Jose, **(2005)**.
[18] T. Acharya and A. Ray, "Image Processing Principles and Applications", John Wiley & Sons, Inc., New Jersey, **(2005)**.

[19] R. Dobrescu, M. Dobrescu, S. Mocanu and D. Popescu, "Medical Images Classification for Skin Cancer Diagnosis Based on Combined Texture and Fractal Analysis", WISEAS Transactions on Biology and Biomedicine, vol. 7, no. 3, **(2010)**, pp. 223-232.
[20] S. G. Wu, F. S. Bao, E. Y. Xu, Y. X. Wang, Y. F. Chang and Q. L. Xiang, "A Leaf Recognition Algorithm for Plant Classification Using Probabilistic Neural Network", IEEE 7th International Symposium on Signal Processing and Information Technology, Cairo, **(2007)**.
[21] T. Deselaers, "Features for Image Retrieval", http://thomas.deselaers.de/publications/papers/deselaers_diploma03.pdf. Aachen – University of Technology, Germany, **(2003)**.

## Authors

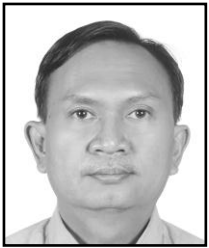

### Abdul Kadir

Received B.Sc. in Electrical Engineering from Gadjah Mada University, Indonesia, in 1987, M. Eng. in Electrical Engineering from Gadjah Mada in 1998, and Master of Management from Gadjah Mada University in 2004. His research interests include image processing, pattern recognition and web-based applications.

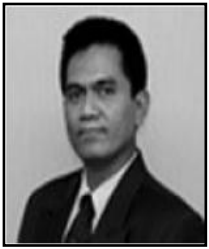

### Lukito Edi Nugroho

Received B.Sc. in Electrical Engineering from Gadjah Mada University, Indonesia, in 1989, M.Sc. from James Cook University of North Queensland in 1994, Ph.D from School of Computer Science and Software Engineering, Monash University, in 2002. His research interests are software engineering, information systems and multimedia.

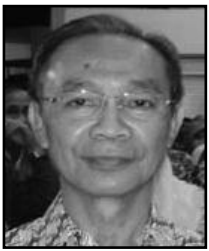

### Adhi Susanto

Professor emeritus at Gadjah Mada University, Indonesia. He received Bachelor in Physics in 1964 from Gadjah Mada University, Indonesia, Master in Electrical Engineering in 1966 from University of California, Davis, USA, and Doctor of Philosophy in 1986 from University of California, Davis, USA. His research interests areas are electronics engineering, signal processing and image processing.

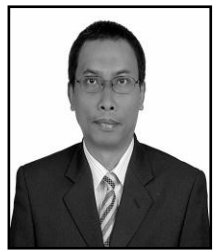

### Paulus Insap Santosa

Obtained his undergraduate degree from Universitas Gadjah Mada, Indonesia, in 1984, master degree from University of Colorado at Boulder in 1991, and doctorate degree from National University of Singapore in 2006. His research interests include human computer interaction and technology in Education.